\documentclass[10pt,twocolumn,letterpaper]{article}

\usepackage{cvpr}              % To produce the CAMERA-READY version
\usepackage{graphicx}
\usepackage{amsmath}
\usepackage{amssymb}
\usepackage{booktabs}
\usepackage{arydshln}
\usepackage{multirow}
\usepackage{etoolbox}
\usepackage{multirow}

\AtBeginEnvironment{tabular}{\small}
\usepackage[pagebackref,breaklinks,colorlinks]{hyperref}

\usepackage[capitalize]{cleveref}
\crefname{section}{Sec.}{Secs.}
\Crefname{section}{Section}{Sections}
\Crefname{table}{Table}{Tables}
\crefname{table}{Tab.}{Tabs.}
\DeclareMathOperator*{\argmin}{arg\,min}

\def\cvprPaperID{11423} % *** Enter the CVPR Paper ID here
\def\confName{CVPR}
\def\confYear{2022}

\begin{document}

%%%%%%%%% TITLE - PLEASE UPDATE
\title{Distill and Collect for Semi-Supervised Temporal Action Segmentation}

\author{Sovan Biswas \and Anthony Rhodes \and Ramesh Manuvinakurike \and Giuseppe Raffa \and Richard Beckwith\\
Intel Labs\\
{\tt\small \{firstname.lastname\}@intel.com}
% For a paper whose authors are all at the same institution,
% omit the following lines up until the closing ``}''.
% Additional authors and addresses can be added with ``\and'',
% just like the second author.
% To save space, use either the email address or home page, not both
% \and
% Second Author\\
% Institution2\\
% First line of institution2 address\\
% {\tt\small secondauthor@i2.org}
}
\maketitle
%%%%%%%%% ABSTRACT
\begin{abstract}
 Recent temporal action segmentation approaches need frame annotations during training to be effective. These annotations are very expensive and time-consuming to obtain. This limits their performances when only limited annotated data is available. In contrast, we can easily collect a large corpus of in-domain unannotated videos by scavenging through the internet. Thus, this paper proposes an approach for the temporal action segmentation task that can simultaneously leverage knowledge from annotated and unannotated video sequences. Our approach uses multi-stream distillation that repeatedly refines and finally combines their frame predictions. Our model also predicts the action order, which is later used as a temporal constraint while estimating frames labels to counter the lack of supervision for unannotated videos. In the end, our evaluation of the proposed approach on two different datasets demonstrates its capability to achieve comparable performance to the full supervision despite limited annotation.
\end{abstract}

%%%%%%%%% Introduction
\section{Introduction}
\label{sec:intro}
Segmenting and understanding a long video sequence is very important for many applications, such as surveillance, intelligent advertisement, \etc It is of high importance to automatic human assistance systems where video sequences tend to have a temporally ordered list of actions. For example, an instructional video of `\textit{making coffee}' tends to have a temporal order of actions such as `\textit{add milk}', `\textit{take cup}', `\textit{stir}', \etc Recently, several approaches have achieved significant success in segmenting activities in videos \cite{lea2017temporal,lei2018temporal,yazan2019mstcn,wang2020boundary,chen2020action}. Despite their success, these approaches are limited by their need for time-consuming and costly-to-obtain frame-level annotation. To overcome these need for frame annotations, many have started to explore approaches to train with weaker supervision in the form of timestamp supervision \cite{li2021temporal,moltisanti2019action,ma2020sf}, transcripts \cite{bojanowski2014weakly,richard2018neuralnetwork,li2019weakly,souri2021fast}, or even sets \cite{fayyaz2020sct,richard2018action,li2020set}. In timestamp supervision, only a single frame of each action segment is annotated. At the same time, the videos are annotated with an ordered list of actions devoid of the starting and ending time in transcript-level supervision. Finally, in set-level supervision, only the set of actions are used in training. These set level annotations lack any order information or indication of the repetition for each action within a video.
\begin{figure}
    \centering
    \includegraphics[scale=0.27]{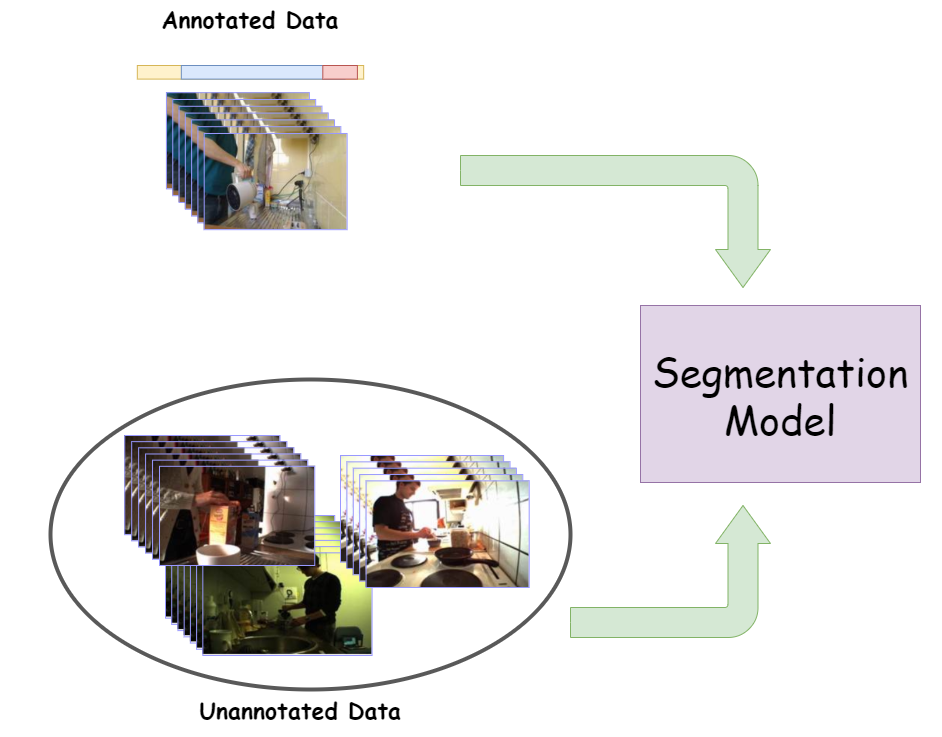}
    \caption{In a semi-supervised setting, we have a combination of a considerable number of unannotated videos and a few annotated video sequences. The aim is to leverage additional information from the \textit{`in-domain'} unannotated video sequences to boost the performance. As annotating frame labels are costly, semi-supervision can help in scaling the training data without additional cost.}
    \label{fig:semisuper}
\end{figure}

On the one hand, weak supervision, such as transcripts or sets, drastically reduces the annotation cost, the overall performance is far from satisfaction as these approaches often fail to correctly align actions in time. On the other hand, an annotator needs to go through each video in detail to generate timestamp supervision. This is a significant bottleneck towards scaling the training data substantially without incurring a high cost. In this paper, we propose a semi supervision approach for the action segmentation task that addresses the limitations of the current level of supervisions. Our semi-supervised data consist of a small amount of detailed frame annotated videos and a large corpus of unannotated videos as shown in Fig \ref{fig:semisuper}. Though at a limited quantity, the use of frame labels provides more substantial supervision than the transcript or set annotation, as it helps to locate and better understand the action variations within a segment. % Compared to timestamp supervision, semi-supervision helps in the quick scaling of training data, which is crucial for further generalization in temporal action segmentation.

Due to the lack of substantial supervision in unannotated videos, leveraging the additional knowledge from these videos can be challenging. A simple approach to solve this problem would be to use the confident frames in the unannotated video based on a threshold as additional frame labels. This, however, raises a question in the form of defining the confidence as classification confidence grows over the training cycle. Further, any early-stage misidentification might propagate the error in later stages - resulting in poor overall performance. Moreover, any model can easily overfit the training data due to few annotations, resulting in false confidence on unannotated videos. Finally, longer video sequences tend to have temporal order across various action segments, which the above approach fails to consider. 

In this work, we propose a different approach where we use predictions from multiple streams of segmentation models to compute temporally ordered frame labels on the unannotated data. Despite limited annotations, the proposed multiple streams generate more accurate and robust predictions by repeated refinement and accumulation. Further, our method also learns to predict an ordered list of actions with a sequence-to-sequence model for the unannotated videos. This predicted ordered list is later used as an alignment constraint to maintain the temporal order while estimating frame labels based on the frame predictions. Any error in the temporal order prediction can significantly impact frame label computing. We, therefore, use beam search to generate multiple ordered candidates to find the optimal matching. This ensures better chances of estimating the correct frame labels for the unannotated data.

Our contributions thus are two folds as follows:
\begin{itemize}
    % \item  We propose to use semi supervision for the temporal action segmentation task, where the goal is to learn the temporal segmentation model with a combination of annotated and unannotated videos. 
    \item We introduce an approach to train a temporal action segmentation model from semi supervision. The approach computes frame labels for an unannotated video aligned to the predicted temporal order of actions.
    \item We propose multi-stream segmentation models that repeatedly refine and accumulate predictions to reduce any noise that may arise from overfitting on the limited training data.
\end{itemize}

%%%%%%%%% Related Work
\section{Related Work}
\label{sec:reltdwrk}

\textbf{Fully Supervised Action Segmentation}: The goal of action recognition \cite{simonyan2014two,carreira2017quo,feichtenhofer2019slowfast} has been to classify short trimmed videos. This is quite different from temporal action segmentation where one requires to capture long-range dependencies in order to classify each frame of the input video. Many approaches in past combined frame-wise classifiers with grammars \cite{vo2014stochastic,pirsiavash2014parsing} or with hidden Markov models (HMMs) \cite{lea2016segmental,kuehne2016end,kuehne2018hybrid}, to capture this dependency. With the recent advancement of neural networks, many recent approaches utilized temporal convolutional networks to capture long-range dependencies for the temporal action segmentation task \cite{lea2017temporal,lei2018temporal}. Despite accurate predictions, these approaches still suffer from an over-segmentation problem. In order to overcome this problem, recent state-of-the-art methods proposes a multi-stage architecture with dilated temporal convolutions \cite{yazan2019mstcn,wang2020boundary,li2020mstcn,huang2020improving,ishikawa2021alleviating,chen2020mixed}. Chen et al. \cite{chen2020action} further enhances the performance by utilizing activity level domain adaptation to compute consistent action labels and reduce over-segmentation.  Yet, these approaches rely on each from annotation resulting in high cost towards real world deployment. On the contrary, in this paper, we perform the temporal action segmentation task in a semi-supervised setup.

\textbf{Weakly Supervised Action Segmentation}: To circumvent the need for expensive annotations of full-supervision, weakly supervised action segmentation has been gaining significant interest recently. Early approaches, in this context, used discriminative clustering based on distinctive action segments to detect and locate temporal action segments using movie scripts \cite{bojanowski2014weakly,duchenne2009automatic}. Despite the apparent presence of temporal ordering in the scripts, these approaches failed to harness them in a meaningful manner. However, Bojanowski et al. \cite{bojanowski2014weakly} were the first to use an ordered list of actions as supervision, thus explicitly modeling the action order to detect and segment actions in a video. Due to the inherent ordering of the actions and the video frames, many researchers recently started addressing the task as an alignment problem between video frames and the transcripts using connectionist temporal classification \cite{huang2016connectionist}, dynamic time warping \cite{chang2019d3tw} or energy-based learning \cite{li2019weakly}. On the other hand, approaches such as \cite{kuehne2017weakly,ding2018weakly,richard2017weakly,kuehne2018hybrid} iteratively generate and refine pseudo ground truth labels for the training. Few approaches, such as  \cite{richard2018neuralnetwork,souri2021fast,souri2021fifa}, use the temporal ordering constraints over-frame predictions to generate the target labels. Lastly, some approaches \cite{fayyaz2020sct,richard2018action,li2020set} tried tackling the action segmentation problem with a much weaker set level supervision without the temporal order. Eventhough these approaches have improved recently, their qualitative performance is far from that of fully supervised approaches due to errors in aligning these actions in time. Bridging the gap, recently \cite{li2021temporal} proposed an approach with timestamp supervision that focused on estimating the action transition in time.  Even though we aim to reduce the annotation cost similar to these approaches, we also want to leverage additional information from unlabeled data. This will help in scaling the training data without any additional cost.

%%%%%%%%% Semi Supervised 
\section{Semi-Supervised Temporal Action Segmentation}
\label{sec:semi-supervised}
The task of the temporal action segmentation task is to predict frame action labels for a given input video. Formally, for a given video of $T$ frames $X = [x_1, \ldots , x_T ]$, the objective is to obtain frame action labels $Y = [y_1, \ldots , y_T ]$. Here $x_t \in \mathbb{R}^D$ and $y_t \in C$ are the frame embedding and action label at time $t$ respectively. $C$ is the set of action classes. The frame labels $Y$ can also be perceived as sequence of $N$ action steps $S = [s_1, \ldots , s_N ]$. Here, $s_n \in C$ is the $n^{th}$ action step. This ordered sequence of steps $S = [s_1, \ldots , s_N ]$ is also known as \textit{video transcript}.  Note the two representations $Y$ and $S$ for are related. 

% For fully supervised action segmentation, we assume all the frame-wise labels $Y$ that are cost and time-intensive to obtain are known during training. This limits incorporating additional training information and thus impacts their scalability. On the other extreme, in weakly supervised action segmentation, all the ordered sequence of action steps (\textit{transcripts}) $S = [s_1, \ldots , s_N ]$ are known. Though transcripts are easier to annotate, the weakly-supervised approaches are limited by their low performance due to the lack of strong frame-wise annotation compared to a full supervised method. Instead, we consider that only a small percentage of the training videos are fully supervised, \ie $Y_s$,  whereas the rest of the training videos, \ie $Y_u$, are neither frame-wise annotated nor provided with action transcripts. Note, the videos are assumed to be `\textit{in-domain}'. The focus is to leverage additional information from un-annotated data to improve the performance of the model. 
For the current semi-supervised formulation, we consider that only a tiny percentage of the training videos ($X_s$) have frame-level supervision ($Y_s$). We scale the training data with additional videos ($X_u$) with neither frame annotations nor action transcripts. Note, both supervised and unsupervised videos are assumed to be `\textit{in-domain}'. The focus is to leverage additional information from unannotated data and improve the overall performance paving the way towards scaling training data with less cost. We now present the proposed method in detail.

%%%%%%%%% Proposed Approach
\section{Proposed Method}
\label{sec:appr}
\begin{figure*}[t]
    \centering
    \includegraphics[scale=0.7]{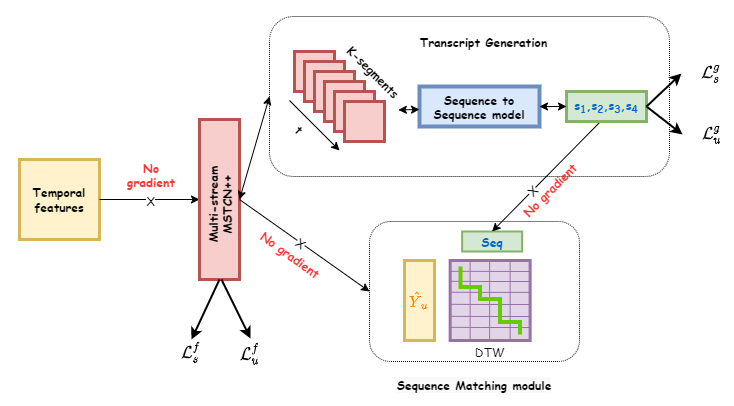}
    \caption{The figure showcases the overview of our method. All bidirectional arrows indicate both the forward and backward pass. During training, we feed a sequence of pre-computed frame features into our proposed multi-stream MSTCN++ to predict frame-level action predictions. For the annotated video sequence, we use these predictions to train our `\textit{Transcription Generation Module}' and `\textit{Multi-Stream MSTCN++}' using $\mathcal{L}_s^g$ and $\mathcal{L}_s^f$ respectively. Here, the \textit{Transcription Generation Module}' predicts the $M$ possible candidate transcripts for an unannotated video sequence. We, then, compute the frame labels $\hat{Y}_u$ for these videos by finding an optimal match in the \textit{Sequence Matching Module}. Lastly, the optimally matched frame labels $\hat{Y}_u$ is also used to further train both `\textit{Transcription Generation Module}' and `\textit{Multi-Stream MSTCN++}' using $\mathcal{L}_u^g$ and $\mathcal{L}_u^f$.}
    \label{fig:overview}
\end{figure*}
Figure \ref{fig:overview} illustrates the overview of the proposed method. Here, the objective of the method is to leverage additional information from unannotated training videos. This leveraging can be done by computing frame labels for those videos. Despite all these videos being `\textit{in-domain}', computing frame labels for an unannotated video is not trivial. Moreover, wrong labels might confuse the model and thus impacting the overall performance. We know that videos have a natural order of actions \eg while making an omelet, \textit{pour oil} always precede \textit{pour egg} action. Thus, our approach focuses on predicting the action order/transcripts for the videos. We argue that the action transcripts should be aligned to their frame action predictions and thus can be used as constraints while computing frame labels for the unannotated videos. Furthermore, noisy or erroneous predictions can impact the label generation despite the transcript constraints. Thus, we also propose multi-stream distillation and collection to repeatedly refine the predictions in a single forward pass in an end-to-end manner.

We first present our temporal action segmentation model based on multi-stream distillation and collection in Section \ref{ssec:dis_acc}. It is then followed by the details of sequence-to-sequence modeling used for the transcript generation model in Section \ref{ssec:seq2seq}. In Section \ref{ssec:similarity}, we describe our sequence matching module to compute the additional frame labels. Finally, we provide the details of all the loss functions in Section \ref{ssec:loss_func}.

\subsection{Distill and Collect: Multi-Stream Predictions}
\label{ssec:dis_acc}
% Our proposed temporal action segmentation model is based on the current state-of-the-art architecture \ie MS-TCN++ \cite{yazan2019mstcn,li2020mstcn} (Section \ref{sssec:mstcn}). 
As shown in Figure \ref{fig:overview}, we generate the additional frame labels and the action transcripts based on the frame action predictions. Thus, any noisy predictions can impact the overall performance. To reduce the noise, we propose a refinement strategy based on multiple streams of segmentation. The objective of the refinement strategy is to repeatedly distill the next segmentation stream and finally collect all the frame predictions across multiple streams.

\subsubsection{MSTCN++: Baseline Model}
\label{sssec:mstcn}
Our proposed approach uses multiple streams of MSTCN++\cite{yazan2019mstcn,li2020mstcn} segmentation models. We first briefly describe the standard MSTCN++ model, which is the current state-of-the-art architecture for action segmentation. The architecture uses pre-computed frame features $X = [x_1, \ldots , x_T ]$ to compute frame labels $\hat{Y}= [\hat{y}_1, \hat{y}_2, \ldots, \hat{y}_t]$. It is performed by multiple stages of the temporal convolutions network. The initial stage is also known as \textit{prediction} or \textit{generation stage}, consisting of a dual dilated layer that combines two temporal convolutions with different dilation factors. It is, then, followed by multiple layers of a single-stage temporal convolution network as known as \textit{ SS-TCN}. A combination of frame-wise cross-entropy loss and smoothing loss \cite{yazan2019mstcn,li2020mstcn} is used to train the segmentation model.

In the current semi-supervised setting, where $(X_s,Y_s)$ and $(X_u,\hat{Y}_u)$ corresponds to annotated and un-annotated video sequence, our frame-wise losses are defined as follows:  
\begin{gather}
    \mathcal{L}_s^f = \frac{1}{T}\sum_{t}{-\log\hat{y}_{t,c}} + \beta * \frac{1}{TC}\sum \hat{\Delta}_{t,c}^2 \label{eq:fl_super}\\
    \mathcal{L}_u^f = \alpha * \frac{1}{T}\sum_{t}{-\log\hat{y}_{t,\hat{c}}} + \beta * \frac{1}{TC}\sum \hat{\Delta}_{t,c}^2 \label{eq:fl_unsuper}
\end{gather}
Here, $\mathcal{L}_s^f$ and $\mathcal{L}_u^f$ are the frame loss over the annotated and unannotated videos, respectively. $\hat{y}_{t,c}$ and $\hat{y}_{t,\hat{c}}$ are the predicted probability for true class $c$ and estimated-label $\hat{c}$, respectively. Same as in \cite{yazan2019mstcn,li2020mstcn}, $\hat{\Delta}_{t,c} = \min(\tau, |\log{\hat{y}}_{t,c}-\log{\hat{y}}_{t-1,c}|)$ and $\tau = 4$ and $\beta=0.15$. $\alpha$ is a weight factor to control the impact of computed-labels as they might be erroneous. 

\subsubsection{Distill - Learn from previous stream:}
\label{sssec:distil}
\begin{figure}[t]
    \centering
    \includegraphics[scale=0.3]{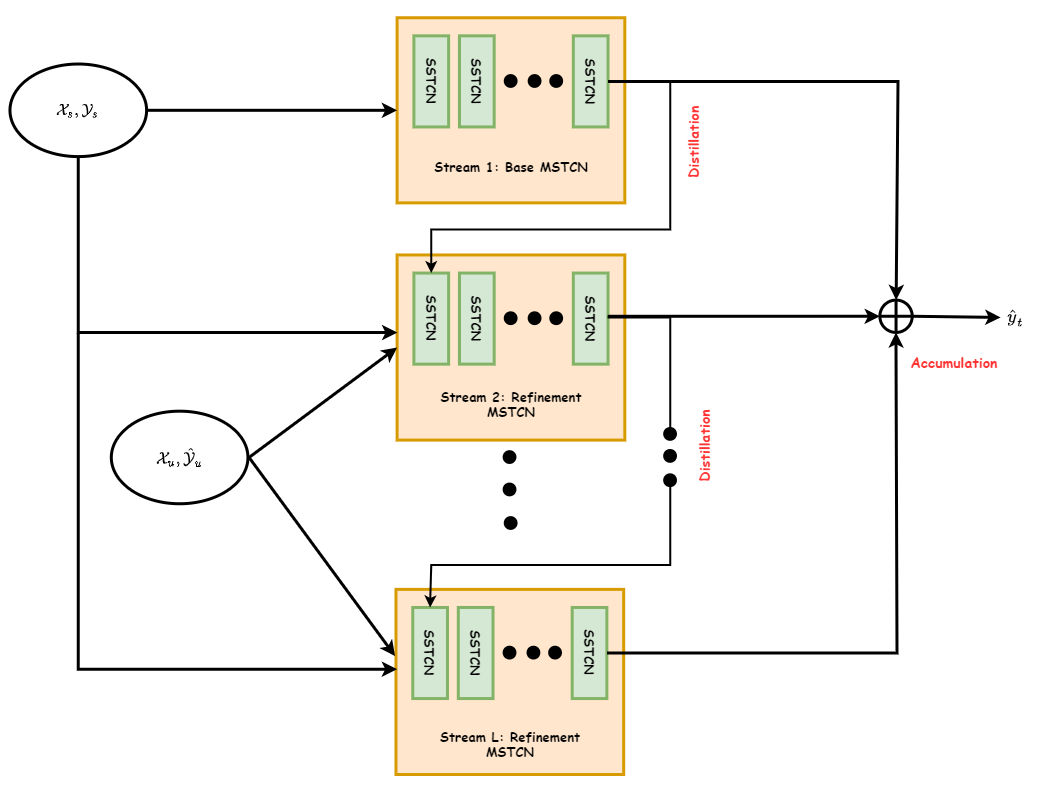}
    \caption{\textit{Distill and Collect - Multiple streams}. The Base MSTCN++ is only trained with ground-truth frame-annotations ($Y_s$). Whereas in the later streams, the predictions of the previous stream are distilled to the first stage of the MSTCN. Note, each refinement stream is additionally trained with both ground truth ($Y_s$) and estimated frame labels ($\hat{Y}_u$).}
    \label{fig:distil}
\end{figure}
In \ref{sssec:mstcn}, we have explained how a single vanilla stream MSTCN++ is trained using ground truth frame labels and estimated labels. However, the estimated frame labels for unannotated videos are computed based on the model predictions, as explained later in Section \ref{ssec:similarity}. Hence, computing robust and generalized frame predictions is imperative to reduce subsequent frame label computation errors. Similar to \cite{tang2017multiple}, we can use multiple streams to computer robust. However, it has some limitations: 1) It is very time-consuming as each stream requires individual computation for label computation; 2) Estimated labels for each stream may vary, which will impact the convergence; 3) It fails to harness the internal structure of MSTCN++.

Instead, we propose distillation across multiple streams as shown in Figure \ref{fig:distil}. Note, the distilled knowledge is only passed to the first stage of the following MSTCN stream instead of the final stage of MSTCN. This enables the following MSTCN to build upon the distilled knowledge. We define our distillation loss ($\mathcal{L}^d$) as follows:
\begin{gather}
    \mathcal{L}^d_l =  \frac{1}{TC}\sum_{t,c}  \min(\tau, |\log{\hat{y}}^l_{t,c}-\log{\hat{y}}^{l-1}_{t,c}|) \\
    \mathcal{L}^d = \sum_{l=2}^{L} \mathcal{L}^d_l \label{eq:distil}
\end{gather}
Here, $l$ and $L$ denotes the index of a stream and the total number of streams, respectively. $\hat{y}^l_{t,c}$ is the predictions of $l^{th}$ stream at time $t$ for a class label $c$. $\tau = 4$. 

Note, the $1^{st}$ stream is only trained with ground truth labels from the annotated videos. Although less generalized, the usage of clean labels at the first stream ensures no-incorrect seed predictions are passed onto the refinement streams. However, in the refinement streams, we additionally use the estimated labels to increase the generalization.

\subsubsection{Collect - Multiple Streams:}
Even though using multi-stream distillation, as explained in \ref{sssec:distil}, helps in refining predictions, ambiguity in predictions still exists. This is especially true at time instants where action transition takes place in the video. We further reduce the noise in predictions by collecting the predictions of all the streams. This collection step is defined as follows:
\begin{equation}
    \log{\hat{y}}_{t,c} = \frac{1}{L}\sum_{l} \log{\hat{y}}^l_{t,c}
    \label{eq:collect}
\end{equation}
$l$ and $L$ denotes the index of a stream and the total number of streams, respectively. $\hat{y}^l_{t,c}$ is the predictions of $l^{th}$ stream at time $t$ for a class label $c$.

Note, our usage of multiple streams increases the size of the overall model in comparison to a single MSTCN++. Post-training, this might be a bottleneck during inference of a test sample where memory is limited. In such cases, the predictions only from the final MSTCN++ stream can be used. Experiments at Section \ref{sssec:tradeoff} showcases the inference performance tradeoff when using the final stream in comparison to the overall model.  
\subsection{Transcript Generation Module:}
\label{ssec:seq2seq}
In this section, we present the details of our transcript generation module that will be used in the computation of frame labels for the unannotated videos in Section \ref{ssec:similarity}. The objective of the transcript generation module is to translate the temporal frame probabilities of a video into its action transcript. To this end, we use sequence-to-sequence model with a bidirectional LSTM encoder to encode the input sequence. This is followed by an LSTM with MLP attention as decoder \cite{bahdanau2014neural}. It is well known that such sequence-to-sequence model is very ineffective in handling long sequences. This is a critical problem as the input video sequence can be significantly longer for such models to handle. However, due to high similarity between consecutive frames, most temporal information is similar and redundant. Thus, we reduce the effective length of the input video sequence by dividing it into $K$ non-overlapping temporal segments. We then temporally `\textit{maxpool}' the class probabilities within these segments to collapse them into $K$ length temporal sequence as shown in Figure \ref{fig:overview}. In this way, we can significantly reduce the temporal length of the video sequence needed for the sequence-to-sequence model to generate action transcripts effectively. 

%We aim to use the transcript as the constraint over-frame prediction to generate additional frame labels on the unannotated data. 
Ideally, we expect the decoder of our transcript generation module to generate the exact action order for those videos. However, in reality, our transcripts may contain a few errors. We use beam-search decoding to generate $M$ possible candidate transcripts for estimating the frame labels, to counteract such errors. Further, if we have additional knowledge about the unannotated video, such as the knowledge of the overall activity, we can add the domain-heuristics during beam-search to generate more relevant transcripts. \eg while preparing coffee, it is highly unlikely that the action \textit{crack egg} is performed. Thus, our decoding can ensure the action \textit{crack egg} is not present in the candidate transcripts.  

We train our Transcript Generation Module with both annotated videos and unannotated videos. The video transcript $S_s$ is easily obtained from the frame labels $Y_s$ to train upon for annotated videos. However, the unannotated videos lack any such information. Instead, we use the best $\hat{S}_u$ among the $M$ candidates after our Sequence Matching Module explained later in Section \ref{ssec:similarity} to train our transcript generation module upon. We define our transcript predictions loss over both annotated and unannotated videos as follows: 
\begin{gather}
    \mathcal{L}_s^g = \frac{1}{N+1}\sum_{n}{-\log\hat{s}_{n,c}} \label{eq:s_trns} \\
    \mathcal{L}_u^g = \alpha * \frac{1}{N+1}\sum_{n}{-\log\hat{s}_{n,\hat{c}}} \label{eq:u_trns}
\end{gather}
Here, $\mathcal{L}_s^g$ and $\mathcal{L}_u^g$ are a cross-entropy loss over ground truth and estimated transcripts respectively. $N$ is the length of the transcript excluding the \textit{EOS} token.  Note, $\hat{s}_{n,c}$ and $\hat{s}_{n,\hat{c}}$ are $n^{th}$ step transcript predictions for annotated step $c$ and estimated step $\hat{c}$ respectively. $\alpha$ is a weight factor to control the impact of estimated transcripts as they might be erroneous. 
\subsection{Sequence Matching Module:}
\label{ssec:similarity}
Given the possible set of transcripts $\mathcal{S}_u$ and the frame probabilities $\hat{y}_{t}$, our aim is to compute the frame-labels ($\hat{Y}_u$) for the unannotated videos. This will enable us to utilize the additional labels from the unannotated videos to train the model better. To this end, we use Dynamic Time Warping \cite{sakoe1978dynamic} to find an optimal alignment between a transcript and the frame predictions. We select the alignment corresponding to the least alignment scores over $M$ candidate transcripts. Note, $|\mathcal{S}_u| = M$. The overall sequence matching module is defined as follows:
\begin{gather}
    Y_S^{*} = \argmin_{Y \in \mathcal{Y}} \langle {Y,C_S} \rangle \label{eq:dtw} \\
    \hat{Y}_u,\hat{S}_u =  \argmin_{Y_S^{*}, S \in \mathcal{S}_u}\langle {Y_S^{*},C_S} \rangle \\
    C_S(t,n) = 1 - \hat{y}_{t,s_n}
\end{gather}
\eqref{eq:dtw} is the equation for the DTW where $C_S$ is the cost matrix between frame probabilities and a candidate transcript $S$. $Y_S^{*}$ is a optimal alignment of the frame predictions with a candidate transcript $S$. $\hat{Y}_u$ and $\hat{S}_u$ are computed frame labels and transcripts based on DTW matching over $M$ candidate transcripts. $\hat{y}_{t,s_n}$ is the action probability at time $t$ and action ($s$) at transcript step $n$.  

\subsection{Loss Functions:}
\label{ssec:loss_func}
As shown in \eqref{eq:fl_super} and \eqref{eq:fl_unsuper}, MSTCN++ combines cross-entropy loss and smoothing loss over both ground-truth and estimated labels. The distillation loss in \eqref{eq:distil} enables the model to refine the predictions across multiple streams repeatedly. Finally, the transcript prediction loss from \eqref{eq:s_trns} and \eqref{eq:u_trns} helps in predicting the action order that is used to generate the estimated labels on the unannotated data. Thus, the overall loss is defined as:
\begin{gather}
    \mathcal{L}_s = \mathcal{L}_s^f + \mathcal{L}_s^g \\
    \mathcal{L}_u = \mathcal{L}_u^f + \mathcal{L}_u^g \\
    \mathcal{L} = \mathcal{L}_s + \mathcal{L}_s + \beta * \mathcal{L}_d \label{eq:overall}
\end{gather}
Here, $\mathcal{L}_s$ and $\mathcal{L}_u$ are the overall losses for annotated videos and unannoated videos respectively. Note, we set $\alpha = 0.3 $ in \eqref{eq:fl_super}  and \eqref{eq:u_trns} and $\beta = 0.15$ for our experiments. 
%%%%%%%%% Experiments
\section{Dataset and Experiments}
\label{sec:appr}
\subsection{Datasets and Metrics}
\label{ssec:datasets}
\textbf{Datasets:} We evaluate our approach on two datasets: Breakfast \cite{kuehne2014language}, and Hollywood Extended \cite{bojanowski2014weakly}.

The \textbf{Breakfast} dataset consist of 1712 videos of breakfast preparing activities with roughly 3.6M frames. Here, the frames are annotated with 48 action classes. The dataset has four splits. We follow the leave-one-split-out strategy for evaluation as explained in \cite{kuehne2014language}. However, to mimic annotated and un-annotated data for the training, we additionally considered leave-one-split annotated, thus effectively reducing the amount of annotation roughly by $1/3$. Note, at any moment, one split is used for testing,  another one split as annotated data, and the rest two are unlabeled data. Our final result is an average of all possible split combinations.

The \textbf{Hollywood extended} dataset contains 937 video sequences taken from Hollywood movies. The videos contain 16 different action classes. We divided that data into training and testing sets of 843 and 94 videos, respectively. This corresponds to seed 5 of the original setting \cite{bojanowski2014weakly}. We, further, divide the training set into three equal splits of 281 videos. To mimic annotated and unannotated data, we assumed only one training split is annotated, and the rest two are unannotated, thus reducing the amount of available annotation by $1/3$. The final result over the test set is the average of all possible combinations of annotated and unannotated training splits.

\textbf{Metrics:} We use the standard metrics for temporal action segmentation such as frame accuracy (MoF), segmental edit distance (Edit), and segmental F1 scores at overlapping thresholds $10\%$, $25\%$ and $50\%$ on the Breakfast dataset. For Hollywood extended, we use the standard non-background frame accuracy (MoF-BG) and intersection over Detection (IoD) as explained in \cite{kuehne2016end}. Additionally, we also report the segmental edit distance (Edit) and segmental F1 scores at overlapping thresholds $10\%$, $25\%$, and $50\%$ for the Hollywood dataset.

\textbf{Baselines:} Our baseline is a standard MSTCN++ \cite{li2020mstcn} when trained only with the annotated data, using the above experimental framework of only one split being annotated. Whereas, MSTCN++ \cite{li2020mstcn} trained on original training data is considered as full-supervised 

\subsection{Implementation Details:}
As explained used MSTCN++ \cite{li2020mstcn} as our base model. We train our model for 12000 steps with Adam optimizer. To minimize the impact of initialization, we compute the frame-wise cross-entropy loss and segment-wise cross-entropy loss on the unannotated data only after the first 2000 steps. We set the learning rate to 0.0005, and the batch size to 3. We use the I3D features per frame for both Breakfast and Hollywood Extended as input to our model, respectively. %We also reduce the impact of the Background class for Hollywood by scaling down the loss over the background class.  

\subsection{Comparison with the Baselines:}
In this section, we compare the proposed approach for semi-supervised action segmentation against a simple baseline (as explained before) that fails to utilize the additional annotated data. The results on the two datasets are shown in Table \ref{tab:baseline}. Despite training on a few annotated data, our approach significantly outperforms these baselines in all the evaluation metrics. It achieves nearly $0.9$ times MoF and $0.87$ times MoF-BG on the full supervised Breakfast and Hollywood extended dataset, respectively.

\begin{table}[t]
\caption{Comparison of the proposed approach with the baseline on both Breakfast and HollywoodExtended Dataset}
\centering
\begin{tabular}{cccccc}
\hline
  \multicolumn{1}{c}{\textbf{Breakfast}}& \multicolumn{3}{c}{F1 = \{$0.1$,$0.25$,$0.5$\}}& \multicolumn{1}{c}{Edit} & \multicolumn{1}{c}{MoF}\\
 \hline
 Baseline & $52.1$ & $46.4$ & $34.4$ & $50.4$ & $54.2$\\
 %Semi-Supervision & $63.9$  & $57.6$ & $43.7$ & $61.8$ & $62.9$ \\
 Semi-Supervision & $\textbf{61.5}$  & $\textbf{55.2}$ & $\textbf{41.7}$ & $\textbf{59.9}$ & $\textbf{60.1}$ \\
 \hdashline
 
 Full-Supervision & $64.1$ & $58.6$  & $45.9$  & $65.6$ & $67.6$ \\
 \hline
 \hline
 \multicolumn{1}{c}{\textbf{Hollywood}}& \multicolumn{3}{c}{F1 = \{$0.1$,$0.25$,$0.5$\}}& \multicolumn{1}{c}{Edit} & \multicolumn{1}{c}{MoF-BG}\\
 \hline
 Baseline & $29.8$ & $25.2$ & $16.1$ & $32.0$ & $35.4$\\
 Semi-Supervision & $\textbf{32.9}$ & $\textbf{28.7}$ & $\textbf{18.8}$ & $\textbf{35.1}$ & $\textbf{38.4}$ \\
 \hdashline
 Full-Supervision & $34.5$ & $31.8$ & $21.4$ & $38.9$ & $44.1$\\
 \hline
\end{tabular}
\label{tab:baseline}
\end{table}

\subsection{Ablation Studies:}
\subsubsection{Impact of Domain Knowledge:}

\begin{table}[h]
\caption{Performance impact of using correct transcripts to estimate labels on the additional videos. \textit{No-Heuristics}, \textit{Domain-Heuristics} and \textit{Mixed-Supervision} corresponds to different level of transcript correctness. The table showcases the result over all the splits of the Breakfast dataset.}
\centering
\begin{tabular}{cccccc}
\hline
  \multicolumn{1}{c}{}& \multicolumn{3}{c}{F1 = \{$0.1$,$0.25$,$0.5$\}}& \multicolumn{1}{c}{Edit} & \multicolumn{1}{c}{MoF}\\
 \hline 
 \textit{No-Heuristics} & $61.5$  & $55.2$ & $41.7$ & $59.9$ & $60.1$\\
 \textit{Domain-Heuristics}  & $63.9$  & $57.6$ & $43.7$ & $61.8$ & $62.9$ \\
 \textit{Mixed-Supervision} & $64.2$  & $58.1$ & $44.4$ & $62.9$ & $63.5$ \\
 \hline
\end{tabular}
\label{tab:domain}
\end{table}

The crux of the proposed algorithm lies in predicting the correct transcripts for the unannotated videos. Any error in the transcripts impacts the sequence matching module and the subsequent estimation of frame labels. Thus, we compare our results (\textit{No-Heuristics}) against the scenario where we have been provided with ground-truth transcripts for the unannotated videos. We denote it as \textit{mixed supervision}. Note, we do not need our \textit{sequence generation module} in such a scenario. Additionally, we experimented with the case where our action predictions are constrained by the \textit{domain-heuristics} of the videos (as explained in \ref{ssec:seq2seq}). Table \ref{tab:domain} showcases the result on the Breakfast dataset with increasing order of transcript correctness. Despite having no prior knowledge about the video content, our approach can still achieve comparable results to mixed supervision.

\subsubsection{Impact of amount of Labelled Data:}
The amount of labelled data impacts the performance of the temporal segmentation. Our approach improves its performance with the increase in labelled data as seen by results on Split 3 (test set) as shown in Table \ref{tab:percentage}. $100\%$ corresponds to the result with full supervision. We also display that our approach achieves nearly $0.77\times$ frame wise accuracy with $\frac{1}{5}$ annotation in comparison to full annotation.  
\begin{table}[h]
\caption{The experiment was performed on Split 3 of the Breakfast dataset. The total training data size remains the same here, including both with frame labels and no labels.}
\centering
\begin{tabular}{cccccc}
\hline
  \multicolumn{1}{c}{$\%$ of labels} &\multicolumn{3}{c}{F1 = \{$0.1$,$0.25$,$0.5$\}}& \multicolumn{1}{c}{Edit} & \multicolumn{1}{c}{MoF}\\
 \hline
 $20\%$  & $59.3$  & $53.4$ & $39.9$ & $59.7$ & $53.6$\\
 $35\%$  & ${61.5}$  & ${55.2}$ & ${42.1}$ & ${61.3}$ & ${60.7}$ \\
 $45\%$  & ${63.3}$  & ${57.9}$ & ${45.0}$ & ${60.1}$ & ${64.6}$ \\
 $100\%$ & ${65.6}$  & ${61.5}$ & ${50.0}$ & ${63.7}$ & ${69.7}$ \\
\hline
\end{tabular}
\label{tab:percentage}
\end{table}

\subsubsection{Impact of Multiple-Stream ($L$):}

The proposed approach accumulates the prediction of multiple streams to have a consistent prediction. This is shown by the quantitative improvement in Table \ref{tab:multi-stream}. In this experiment, we utilize the Breakfast dataset with domain heuristics. The results improve from a frame accuracy of $57.9\%$ to $62.9\%$ when the number of streams increases from $2$ to $4$. We also experimented with full-supervised models and usage of multiple streams. Our experiments improve the frame accuracy from $67.6\%$ to $70.1\%$ when the number of streams is increased from $1$ to $4$. Note, there is no change in each stream of the MSTCN++ model during full-supervision apart from the use of proposed distillation (\ref{eq:distil}) and collection (\ref{eq:collect}).

\begin{table}[h]
\caption{Performance impact of using multiple streams on the split of Breakfast dataset. $L$ denotes the number of streams in the multi-stream architecture. Note, this experiment used domain-heuristics during the semi-supervised training.}
\begin{tabular}{cccccc}
\hline
  \multicolumn{1}{c}{Semi-Supervision}& \multicolumn{3}{c}{F1 = \{$0.1$,$0.25$,$0.5$\}}& \multicolumn{1}{c}{Edit} & \multicolumn{1}{c}{MoF}\\
 \hline 
 $L=2$ & $57.3$ & $51.2$ & $38.2$ & $56.0$ & $58.5$\\
 $L=3$ & $62.7$  & $56.2$ & $42.7$ & $61.0$ & $61.5$ \\
 $L=4$ & $\textbf{63.9}$  & $\textbf{57.6}$ & $\textbf{43.7}$ & $\textbf{61.8}$ & $\textbf{62.9}$ \\
 \hline
 \hline
 \multicolumn{1}{c}{Full-Supervision}& \multicolumn{3}{c}{F1 = \{$0.1$,$0.25$,$0.5$\}}& \multicolumn{1}{c}{Edit} & \multicolumn{1}{c}{MoF}\\
 \hline
 $L=1$ & $64.1$ & $58.6$  & $45.9$  & $65.6$ & $67.6$\\
 $L=4$ & $\textbf{67.1}$ & $\textbf{61.9}$  & $\textbf{50.4}$  & $\textbf{64.2}$ & $\textbf{70.1}$ \\
 \hline
\end{tabular}
\label{tab:multi-stream}
\end{table}

\subsubsection{Impact of Distillation vs Collection:}

\begin{table}[h]
\caption{Performance impact of distillation and collection step. The experiments was performed only on the split 1 of the Breakfast dataset.}
\centering
\begin{tabular}{cccccc}
\hline
  \multicolumn{1}{c}{Approach}& \multicolumn{3}{c}{F1 = \{$0.1$,$0.25$,$0.5$\}}& \multicolumn{1}{c}{Edit} & \multicolumn{1}{c}{MoF}\\
 \hline 
 Baseline & $56.1$  & $50.5$ & $38.3$ & $54.6$ & $56.7$\\
 \textit{Only Distil} & $58.3$  & $52.4$ & $38.9$ & $59.7$ & $58.9$\\
 \textit{Only Collect} & $62.6$  & $56.4$ & $41.8$ & $61.1$ & $62.1$ \\
 \textit{Both} & $\textbf{63.4}$ & $\textbf{57.3}$ & $\textbf{43.5}$ & $\textbf{61.6}$ & $\textbf{63.0}$ \\
 \hline
\end{tabular}
\label{tab:dis_col}
\end{table}

Our proposed approach uses a combination of distillation and collection of predictions to reduce noise and increase robustness. Though, in theory, both contribute towards better prediction, the question remains with their overall impact. In Table \ref{tab:dis_col}, we disentangle each of the processes individually. In the experiment, \textit{Only Distil} uses only distillation without the collection step \eqref{eq:collect}. In this case, the prediction of the final stream is used to compute the labels and during inferencing, both. \textit{Only Collect} uses multiple stream without the distillation loss \eqref{eq:distil}. Though distillation improves the result compared to the baseline, the experiment showcases that the collection of predictions over multiple streams contributes more towards the overall approach. 

\subsubsection{Impact of $M$ candidates:}

Beam-search decoding enables our approach to generate multiple transcript candidates. Table \ref{tab:beam} showcases the impact of various candidates on the overall performance. Using a value of $M=5$ increases the result slightly compared to just greedy decoding ($M=1$) for a transcript. Note, the computation cost of the sequence matching module increases with an increase in value of $M$. Thus, we set $M=5$ for all our experiments, which enables us to rectify any greedy decoding errors and retain tractable computation costs.

\begin{table}[h]
\caption{Performance impact due to $M$ candidates. The experiment was performed on all the splits of the Breakfast dataset.}
\centering
\begin{tabular}{cccccc}
\hline
  \multicolumn{1}{c}{}& \multicolumn{3}{c}{F1 = \{$0.1$,$0.25$,$0.5$\}}& \multicolumn{1}{c}{Edit} & \multicolumn{1}{c}{MoF}\\
 \hline 
 $M=1$ & $61.2$  & $54.9$ & $40.8$ & $58.9$ & $59.9$\\
 $M=5$ & $\textbf{61.5}$  & $\textbf{55.2}$ & $\textbf{41.7}$ & $\textbf{59.9}$ & $\textbf{60.1}$ \\
\hline
\end{tabular}
\label{tab:beam}
\end{table}

\subsubsection{Performance tradeoff: Using Final Stream vs Overall Model:}
\label{sssec:tradeoff}
\begin{table}[h]
\caption{Performance tradeoff when using final stream vs overall model. Here, \textit{Baseline} is the model trained only on the annotated data. The experiment was performed on all the splits of the Breakfast dataset.}
\centering
\begin{tabular}{cccccc}
\hline
  \multicolumn{1}{c}{Model Size}& \multicolumn{3}{c}{F1 = \{$0.1$,$0.25$,$0.5$\}}& \multicolumn{1}{c}{Edit} & \multicolumn{1}{c}{MoF}\\
 \hline
 \textit{Baseline} & $52.1$ & $46.4$ & $34.4$ & $50.4$ & $54.2$\\
 \textit{Last Stream} & $59.5$  & $53.1$ & $39.7$ & $58.4$ & $58.2$\\
 \textit{Overall} & $\textbf{61.5}$  & $\textbf{55.2}$ & $\textbf{41.7}$ & $\textbf{59.9}$ & $\textbf{60.1}$ \\
 \hline
\end{tabular}
\label{tab:tradeoff}
\end{table}
As stated earlier, we collect the predictions of all the streams to compute the final prediction. This might be a bottleneck during inference with limited memory. A simple tradeoff uses only the final stream instead of the whole trained model and ignores the collection step. Table \ref{tab:tradeoff} showcases the performance tradeoff in such cases. Despite the drop in performance when using only the final stream, it is still far better than the Baseline model, where additional training data was ignored. 

\section{Ablation-Studies: Impact of $\alpha$}
In Eq. \eqref{eq:fl_unsuper} and Eq. \eqref{eq:u_trns},  we use a weight factor $\alpha$ to control the errors in the estimated transcripts from impacting the overall model. If the estimated transcripts were accurate and $\alpha = 1$, this proposed semi-supervised approach would be equivalent to full-supervision. Table \ref{tab:alpha} showcases the impact of the $\alpha$ on the overall performance. Increasing the value of $\alpha$ impacts the performance negatively as the model is forced to learn the errors with higher confidence. On the other hand, setting $\alpha$ to a lower value enables the approach to balance the impact of the transcripts error towards the overall frame label prediction. Thus, we set our value to $\alpha=0.3$ for all our experiments.

\begin{table}[h]
\caption{Performance impact of $\alpha$ on the overall performance. The table showcases the result over all the splits of the Hollywoodextended dataset.}
\centering
\begin{tabular}{ccccccc}
 \hline
 \multicolumn{1}{c}{\textbf{$\alpha$}}& \multicolumn{3}{c}{F1 = \{$0.1$,$0.25$,$0.5$\}}& \multicolumn{1}{c}{Edit} & \multicolumn{1}{c}{MoF-BG}&\multicolumn{1}{c}{IoD}\\
 \hline
 Baseline & $29.8$ & $25.2$ & $16.1$ & ${32.0}$ & $35.4$ & -\\
\hdashline
 $0.1$ & $32.2$ & $28.4$ & $17.9$ & $\textbf{35.7}$ & $37.8$ & $\textbf{52.1}$\\
 $0.3$ & $\textbf{32.9}$ & $\textbf{28.7}$ & $\textbf{18.8}$ & ${35.1}$ & $\textbf{38.4}$ & $51.4$ \\
 $1.0$ & $28.5$ & $24.8$ & $15.1$ & $33.9$ & $35.1$ & $49.7$\\
 \hline
\end{tabular}
\label{tab:alpha}
\end{table}

\AtBeginEnvironment{tabular}{\scriptsize}
\begin{table}[ht]
% \begin{minipage}[b]{0.5\linewidth}
\caption{Comparison of the proposed approach with the state-of-the-art on the Breakfast dataset. $*$ indicates the authors have used more $\%$ of annotation in comparison.}
\begin{tabular}{ccccccc}
\hline
  \multicolumn{1}{c}{}&\multicolumn{1}{c}{Method}& \multicolumn{3}{c}{F1 = \{$0.1$,$0.25$,$0.5$\}}& \multicolumn{1}{c}{Edit} & \multicolumn{1}{c}{MoF}\\
 \hline
 \multirow{4}{*}{Full} & MSTCN\cite{yazan2019mstcn} & $52.6$ & $48.1$ & $37.9$ & $61.7$ & $66.3$\\
 & MSTCN++\cite{li2020mstcn} & $64.1$  & $58.6$ & $45.9$ & $65.6$ & $67.6$\\
 % & Proposed & $67.1$ & ${61.9}$  & ${50.4}$  & ${64.2}$ & ${70.1}$ \\
 & BCN\cite{wang2020boundary} & $68.7$  & $65.5$ & $55.0$ & $66.2$ & $70.4$\\
 & ASRF\cite{ishikawa2021alleviating} & $74.3$  & $68.9$ & $56.1$ & $72.4$ & $67.6$\\
 \hline
 \multirow{1}{*}{{Timestamps}} & \tiny{MSTCN+Timestamps}\cite{li2021temporal} & $70.5$ & $63.6$ & $47.4$ & $69.9$ & $64.1$\\
 % & Proposed & $67.1$ & ${61.9}$  & ${50.4}$  & ${64.2}$ & ${70.1}$ \\
 \hline
 \multirow{2}{*}{Semi} & {HMM-RNN}$^{*}$\cite{kuehne2018hybrid} & - & - & - & - & $61.3$\\
 & {Ours} & $\textbf{63.9}$ & $\textbf{57.6}$ & $\textbf{43.7}$ & $\textbf{61.8}$ & $\textbf{62.9}$\\
 % & Proposed & $67.1$ & ${61.9}$  & ${50.4}$  & ${64.2}$ & ${70.1}$ \\
 \hline
 \multirow{4}{*}{Transcripts} & {CDFL}\cite{li2019weakly} & - & - & - & - & $50.2$\\
 & {Mucon}\cite{souri2021fast} & - & - & - & - & $47.1$\\
 & {D3TW}\cite{chang2019d3tw} & - & - & - & - & $45.7$\\
 & {NN-Viterbi}\cite{richard2018neuralnetwork} & - & - & - & - & $43.0$\\
 \hline
 \multirow{4}{*}{Sets} & {SCT}\cite{fayyaz2020sct} & - & - & - & - & $30.4$\\
 & {SCV}\cite{li2020set} & - & - & - & - & $30.2$\\
 & {Action-Sets}\cite{richard2018action} & - & - & - & - & $23.3$\\
 \hline
\end{tabular}
\label{tab:sotaBF}
% \end{minipage}
\end{table}
\AtBeginEnvironment{tabular}{\small}

% \begin{table}[h]
% \caption{Performance impact due to $M$ candidates. The experiment was performed on all the splits of the Breakfast dataset.}
% \centering
% \begin{tabular}{cccccc}
% \hline
%   \multicolumn{1}{c}{}& \multicolumn{3}{c}{F1 = \{$0.1$,$0.25$,$0.5$\}}& \multicolumn{1}{c}{Edit} & \multicolumn{1}{c}{MoF}\\
%  \hline 
%  $M=1$ & $61.2$  & $54.9$ & $40.8$ & $58.9$ & $59.9$\\
%  $M=5$ & $\textbf{61.5}$  & $\textbf{55.2}$ & $\textbf{41.7}$ & $\textbf{59.9}$ & $\textbf{60.1}$ \\
% \hline
% \end{tabular}
% \label{tab:beam}
% \end{table}

% \setlength\doublerulesep{0.5mm}
\begin{table}[ht]\centering
% \begin{minipage}[b]{0.5\linewidth}
\caption{Comparison of the proposed approach with the state-of-the-art on the Hollywood dataset.}
\begin{tabular}{cccc}
\hline
   \multicolumn{1}{c}{}&{Method}& MoF-BG & IoD\\
 \hline
 \multirow{4}{*}{Transcripts} & {CDFL}\cite{li2019weakly} & $40.6$ & $52.9$\\
 & {Mucon}\cite{souri2021fast} & $41.6$ & $52.3$\\
 & {D3TW}\cite{chang2019d3tw} & $33.6$ & $50.9$\\
 & {NN-Viterbi}\cite{richard2018neuralnetwork} & - & $48.7$\\
 \hdashline
  \multirow{2}{*}{Semi} & HMM-RNN\cite{kuehne2018hybrid} & -  & $13.7$\\
 & Ours & $\textbf{38.4}$ & $\textbf{51.4}$ \\
 \hline
\end{tabular}
\label{tab:sotaHE}
% \end{minipage}
\end{table}

\subsection{Comparison to State-of-the-art}
In this section, we compare our approach with semi-supervised annotation against the recent state-of-the-art approaches. To the best of our knowledge, semi supervision has not been studied extensively. Results for the Breakfast, and Hollywood extended datasets are shown in Table \ref{tab:sotaBF} and Table \ref{tab:sotaHE}, respectively. Our approach is comparable to timestamp supervision and full-supervision despite using roughly ${1/3}^{rd}$ of training data on Breakfast dataset. This demonstrates the capability of our approach to scaling quickly when additional `\textit{in-domain}' data is available. We also compare our approach with the weakly supervised setup where the action order in the form of transcripts is only provided. Due to the availability of detailed frame annotations for a few videos, the model performs significantly better than any transcripts based approach on Breakfast dataset. On Hollywood extended, however, our approach is on par with few state-of-the-art transcript based approaches.

%%%%%%%%% Conclusion
\section{Conclusion}
\label{sec:concl}
In this paper, we proposed an approach that leverages additional knowledge from `in-domain' unannotated videos to train a temporal action segmentation model when a limited annotation is available. Our approach is based on multi-stream distillation that repeatedly refines and finally combines their frame predictions. We further predicted the action order that is used as an order constraint while computing the frame labels of unannotated videos. Our experiments on two different action datasets showcase comparable performance to full-supervised approaches despite limited annotation. Additionally, we also demonstrate the impact of the proposed distillation and accumulation on full supervision, pointing towards its potential across various levels of supervision.
{\small
\bibliographystyle{ieee_fullname}
\bibliography{egbib}
}

\end{document}